# Epistemic Norms for AI Safety and Alignment Research

Keivan Navaie
Lancaster University, UK (k.navaie@lancaster.ac.uk)

## Abstract

Mainstream AI research emphasises capability growth and tolerates low failure rates when average-case performance is high. AI safety and alignment research has a different mission: to ensure that catastrophic failures never occur, under sparse evidence, adversarial dynamics, and fat-tailed risk. We argue that the two domains differ along two analytically independent axes — *capability profile* (demonstrating the absence of hazardous behaviours versus the presence of positive capabilities) and *risk profile* (bounding worst-case outcomes under fat-tailed uncertainty versus optimising average-case performance) — and that mainstream epistemic practices are inadequate on both. Building on a structured synthesis grounded in a preregistered bibliometric baseline, we identify five cross-cutting gap dimensions in current alignment research, including the near-absence of institutionalised independent verification. To address these gaps we propose ECAISA, an Epistemic Code for AI Safety and Alignment comprising eight principles, a three-level scoring rubric, a four-level disclosure ladder that reconciles transparency with information-hazard and commercial-confidentiality constraints, a tiered applicability scheme, an infohazard adjudication procedure, and seven anti-gaming mechanisms. ECAISA does not certify that any AI system is safe; it constrains how safety-relevant research claims are documented, checked, and relied upon, with *auditability rather than certification* as its governance target. A retrospective rubric audit ($\kappa = 0.79$) demonstrates instrument feasibility; a four-stage validation roadmap is proposed.

## 1. Introduction

Epistemic norms tell a research community when a claim is "knowledge-grade," that is, justified enough to count as reliable knowledge. Under classical reliabilist theories, a belief qualifies as knowledge only if the inferential process that produced it is truth-conducive and reliably leading from true premises to true conclusions (Armstrong 1973). Robert K. Merton articulated this ideal in four operational norms: communalism (open sharing of findings), universalism (method over authority), disinterestedness, and organised scepticism (Merton 1942/1973). In practice, an epistemic norm is a concrete standard that tells researchers when to accept, share, or challenge a claim; examples include "findings must be independently replicable", "claims must survive adversarial challenge", and "hypotheses must be preregistered". Such norms are the operational counterpart to the abstract ideals of communalism and scepticism.

Mainstream AI has, to a meaningful degree, operationalised these epistemic ideals by anchoring progress to shared public benchmarks and leaderboards, instituting reproducibility checklists and artifact-review tracks, and normalizing code/data release alongside rigorous (if imperfect) peer review. ImageNet, GLUE, and SuperGLUE exemplify community benchmarks that make results publicly comparable (Russakovsky et al. 2015; Wang et al. 2019, 2020). NeurIPS's reproducibility program and related efforts formalize reporting standards and require code, data, and hyperparameters for audit (Pineau et al. 2021; Gundersen & Kjensmo 2018). And while incentives remain uneven, norms of releasing implementation details and subjecting work to conference/journal peer review have helped sustain a baseline of credibility (Raff 2019; Langley 2019).

AI safety and alignment research poses a fundamentally different goal and risk profile: its mission is preventive to ensure that certain harmful outcomes (e.g., loss of human control over advanced AI) never occur whereas mainstream AI primarily seeks to create capabilities and measures success by positive task performance. The concept of alignment as a preventive, risk centred- agenda separate from purely capability -driven AI development was first articulated by Bostrom (2011, 2014) and later elaborated by Russell, Dewey & Tegmark (2015) and Amodei et al. (2016).

We use “AI safety” to denote the broad sociotechnical effort to prevent and mitigate harms from AI systems across their lifecycle, spanning technical controls, organisational processes, and governance standards such as the NIST AI RMF, ISO/IEC 42001, and the EU AI Act. By contrast, “AI alignment” names the technical problem of ensuring that highly capable systems’ objectives and behaviours remain acceptably compatible with human values under worst-case, adversarial, and distribution-shift conditions. Alignment therefore carries a preventive mission (“prove the negative”) and a fat-tailed risk profile, requiring very low tolerance for catastrophic error and privileging adversarial verification over average-case performance gains. Throughout, we treat alignment as a core but not exhaustive component of the wider AI safety agenda; when we say “AI safety and alignment”, we refer to this coupled scope.

In scientific terms, alignment is an evidence-of-absence problem: we must justify, to very high confidence, that no unacceptable failure will occur, not merely show a positive capability. Because many plausible harms (e.g., an accidental existential catastrophe) would be irreversible, the burden of proof is sharply asymmetric and precautionary. The challenge is compounded by the possibility that advanced AI systems behave adversarially or deceptively, strategically masking misalignment during evaluation, which makes some failures intrinsically hard to foresee. Evidence for such adversarial/deceptive dynamics includes work on *deceptive alignment* (Hubinger et al. 2019), Alignment Faking (Greenblatt et al. 2024) and adversarial examples (Goodfellow, Shlens & Szegedy 2015), with further catalogues of hard-to-anticipate safety failures and goal mis-generalization in Amodei et al. (2016) and Langosco et al. (2022).

Because we cannot ethically “test to failure” when the downside could be civilisation-scale harm, alignment research must proceed under sparse, indirect evidence, deep epistemic uncertainty, and fat-tailed risk where worst-case losses dwarf average performance. Any claim of safety or alignment therefore demands evidence far stronger than the metrics that suffice for ordinary ML benchmarks evidence organised as a safety case rather than a performance plot. This stricter standard echoes a growing body of work: Shevlane et al. (2023) call for “very strong assurance” in alignment evaluations; Leveson (2012) distinguishes rigorous safety assurance from mere reliability; Amodei et al. (2016) advocate accident-oriented validation for AI systems; and Bloomfield and Rushby (2024) along with Hilton et al. (2025) adapt high-assurance safety-case methodology specifically to AI.

Furthermore, in most AI fields a 99.9% success rate might be considered excellent; but in alignment, even a 0.1% risk of failure could hide an existential catastrophe. Standard AI research incentives (novelty, rapid publication) tend to undervalue the extreme tail risks of AI, since they reward average-case improvements. We therefore argue that AI safety research instead requires an epistemic regime of extreme rigour: full transparency of methods, systematic adversarial testing, and very conservative inference.

This mirrors the ethos of high-reliability engineering domains such as aviation, medicine, nuclear power, and high-assurance cyber security, where deployment is authorised only after stringent demonstrations of safety or resilience. Commercial aviation, for example, treats catastrophic-failure probabilities below $10^{-9}$ per flight-hour as a design-assurance target (FAA Advisory Circular 25.1309-1B); critical-infrastructure cyber security demands formal verification, exhaustive penetration testing, and zero-day resilience before operational release. The $10^{-9}$ figure serves here as a motivating analogy, not as a computational target for alignment: alignment lacks the actuarial base-rate data, well-defined failure modes, and stationary dynamics that make a specific numerical threshold well-posed. The principle we take from these precedents is evidence-proportionality, namely that the stringency of the evidence required for a safety claim should scale with the severity and irreversibility of the claim's potential failures, not a specific probability target. By this principle, every safety claim in AI alignment should be scrutinised as though it were false, and even seemingly negligible risks cannot be dismissed without justification.

In this paper, we adopt a mixed methodology that combines conceptual analysis with comparative benchmarking. We begin by identifying the key asymmetries between alignment and conventional AI research and then conduct a gap analysis to pinpoint where mainstream epistemic norms fall short for alignment's needs. We then draw on lessons from safety-critical fields and from the open science and metascience literature to inform this analysis. Based on these insights, we formulate a concrete proposal: a codified Epistemic Code for AI Safety and Alignment (ECAISA) consisting of eight core principles tailored to the alignment community.

Our core contributions are: (i) a gap analysis between mainstream AI research norms and alignment requirements, grounded in a preregistered bibliometric baseline (§3.1); (ii) the ECAISA code of eight epistemic principles, supported by a three-level scoring rubric for application by venues, funders, and laboratories (§5, Appendix A); (iii) a governance crosswalk that maps each principle to NIST AI RMF 1.0, ISO/IEC 42001, and the EU GPAI Code of Practice, with explicit statement of what ECAISA adds (§6.1, Table 7); (iv) a retrospective rubric audit of eight alignment papers ($\kappa = 0.79$) demonstrating instrument feasibility, together with a four-stage validation roadmap (§7); and (v) an analysis of adoption dynamics that treats unilateral adoption as a coordination problem and identifies three institutional paths past first-mover disadvantage (§7.2).

In the remainder of this paper, we elaborate each of these contributions, beginning with the fundamental epistemic differences that set alignment apart.

# 2. Epistemic Asymmetries Between Mainstream AI and Alignment

Two asymmetries distinguish safety and alignment research from mainstream AI. We state each along a single axis, so that the two are analytically distinct and the gap analysis of §3 does not merely restate them. The first is an asymmetry of *capability profile*: mainstream AI demonstrates what a system can do, whereas alignment must demonstrate what a system cannot be made to do. The second is an asymmetry of *risk profile*: mainstream AI optimises average-case performance, whereas alignment must bound worst-case outcomes under fat-tailed, partly adversarial uncertainty. These two axes are related but independent; a study can be strong on one and weak on the other.

## 2.1 Capability profile: demonstrating absence versus presence

Mainstream AI demonstrates positive capabilities: higher accuracy, more fluent generation, stronger reward. The object of demonstration is what the system can do. Alignment inverts this: the object of demonstration is what the system must not do, even when pushed. That is, alignment must provide evidence for the absence of hazardous behaviours, including behaviours that adversarial users, distribution shifts, or the system itself may elicit under conditions not present in the evaluation. Safety-critical engineering offers the relevant analogy. Avionics and medical-device certification do not rely on strong mean performance; they require demonstrations that specific failure modes will not occur at unacceptable rates across operating conditions (Littlewood and Strigini 1993; Leveson 2012; RTCA DO-178C 2011; ISO 14971:2019; FAA AC 25.1309-1B).

The stance that follows is Popperian (Popper 1959; Mayo 2018): safety claims are treated as conjectures that must survive severe tests, not as summaries of observed successes. In practice this means systematic adversarial testing, scenario analysis, and formal verification applied to candidate failure modes (Amodei et al. 2016; Hubinger et al. 2019; Shevlane et al. 2023; Madry et al. 2018; Katz et al. 2017). A 99.9% benchmark score is less informative than the 0.1% of cases that failed, because it is the failure set that the safety claim must rule out. This asymmetry is about the object of demonstration; it is orthogonal to how the probability of failure is distributed, which is the subject of the second asymmetry below.

## 2.2 Risk profile: worst-case under fat-tailed uncertainty

Mainstream AI evaluation optimises and reports average-case performance. Alignment must instead bound the probability and magnitude of worst-case outcomes under distributions that are fat-tailed, sparsely sampled, and partly adversarial. The distinction matters because a system can post excellent average performance while concealing rare but severe failures (Cirillo and Taleb 2020; Bostrom 2011; Shevlane et al. 2023). Where the loss distribution has a long tail, mean performance is not a summary of the object of interest; the tail is.

The analogy to high-reliability engineering is instructive here. Nuclear regulators do not judge safety by expected outcomes; they require defence in depth, redundancy, and continuous auditing of the assumptions behind the safety case, and they quantify residual risk using explicit uncertainty analysis rather than point estimates (IAEA SSR-2/1 Rev.1, 2016; NRC Regulatory Guide 1.174 Rev.3, 2018; Littlewood and Strigini 1993; Leveson 2012; NRC NUREG-1855 Rev.1, 2017). Alignment inherits this posture in full, and adds one feature that high-reliability engineering largely does not face: the adversarial component of the risk profile. Advanced models may mask unsafe behaviour during evaluation, so an apparent pass may itself be evidence of deceptive alignment (Hubinger et al. 2019). This requires continuous red-teaming, scepticism toward tests that appear to pass, and formal safety-assurance methodology (Bloomfield and Bishop 2010; ISO/IEC 15026-2:2011; Shevlane et al. 2023; Perez et al. 2022; Ganguli et al. 2022).

The risk profile is about the probabilistic target of evidence; it is independent of, and orthogonal to, the capability profile set out above. A study may optimise for worst-case performance and still commit the mainstream-capability framing by reporting only successes (weak on capability profile). Likewise, a study may adopt the absence-of-hazard framing but report only mean-adversarial performance (weak on risk profile). Treating the two as independent lets the gap analysis in §3 diagnose failure modes along one axis without covertly invoking the other.

## 2.3 Safety as a sociotechnical property of the research record

A clarification of scope is necessary before proceeding to the gap analysis. Safety is not a property of technical systems in isolation but of how those systems are developed, evaluated, and deployed within social and institutional contexts (Leveson 2012; Wieringa 2020). ECAISA operates at a level consistent with this framing. It specifies the evidentiary practices that make a research-phase safety claim auditable, that is, inspectable and contestable by independent reviewers, funders, oversight bodies, and affected communities. The object of ECAISA's requirements is the research record: the set of artefacts, disclosures, and documented practices that constitute the evidence base for a safety-relevant claim. This is itself a sociotechnical object: produced within institutional contexts, shaped by incentive structures, and consumed by decision-makers who rely on it to authorise deployment, allocate funding, or set policy. The five epistemic gaps identified in §3 are therefore gaps in the social production of evidence, not merely technical measurement failures; ECAISA's principles are, in this sense, governance interventions as much as epistemic ones.

We also acknowledge the epistemological limits of any assurance framework for open-ended systems (Narayanan and Kapoor 2023). Because the input and output spaces of capable AI systems are effectively unbounded, no evaluation programme can close the denominator of the failure probability; comprehensive safety assurance for general-purpose systems is not achievable by any finite set of tests. ECAISA does not attempt to achieve it. The appropriate analogy is clinical research: randomised trials cannot enumerate all patient contexts or eliminate residual risk, yet structured evidentiary standards demonstrably improve the contestability and reliability of therapeutic claims relative to unstructured practice. ECAISA makes the same more modest claim: a research record produced under P1–P8 is more auditable, more contestable, and harder to misuse as a compliance proxy than one produced under current norms, regardless of whether it achieves certainty.

These asymmetries mean that alignment researchers cannot simply import the evaluation norms of mainstream AI. Instead, they must elevate standards of evidence and rigor to account for the higher stakes and trickier epistemics.

We next analyse specific dimensions along which mainstream AI research practices versus alignment needs diverge and identify gaps that any adequate epistemic framework for AI safety must address.

# 3. Gap Analysis: Mainstream Norms *vs.* Alignment Needs

The two asymmetries of §2 describe what alignment must demonstrate. They do not by themselves specify where mainstream research practice falls short. To go from the asymmetries to diagnostic leverage, we introduce five cross-cutting dimensions along which mainstream norms and alignment needs diverge. Each dimension is an instrument that may surface failure modes on either asymmetry: the same gap (for example, weak verification) can allow a capability-profile failure to go undetected and allow a risk-profile failure to go unquantified. The five dimensions are therefore not further instances of the two asymmetries but the observable channels through which both asymmetries are violated in practice.

Mainstream AI research is built around average-case optimisation (benchmark scores, mean accuracy, expected reward) and is typically accompanied by selective post-hoc artefact sharing, limited independent verification, shallow uncertainty reporting, and weak incentives to reward negative results or adversarial critique (Russakovsky et al. 2015; Wang et al. 2019; Gundersen and Kjensmo 2018; Henderson et al. 2018; Raff 2019; Pineau et al. 2021; Munafò et al. 2017). Alignment has to invert these priorities: optimise for worst-case (tail) safety, demand radical up-front openness, invite adversarial multi-team replication and stress-testing, embrace multi-layered uncertainty accounting, and foster institutionalised scepticism (Bostrom 2014; Shevlane et al. 2023; Hubinger et al. 2019; Amodei et al. 2016; Madry et al. 2018; Leveson 2012). The five dimensions below are the places where this inversion either happens or fails to happen.

1. **What is being optimised against** (failure-mode targeting vs. mean-performance targeting)
2. **Visibility of evidence** (transparency)
3. **Trustworthiness of evidence** (verification & replication)
4. **Treatment of unknowns** (uncertainty accounting)
5. **Enforcement culture** (incentives and norms for rigour).

Each dimension corresponds to well-documented failure modes in other high-reliability fields. Mapping these gaps yields a focused rubric for diagnosing where conventional AI research practices break down when transplanted into the alignment domain.

## 3.1 Bibliometric baseline (preregistered protocol)

The five gap dimensions above are derived from a structured comparative synthesis of safety-engineering literature, meta-scientific work on reproducibility and selective reporting, and published critiques of AI evaluation practice (Henderson et al. 2018; Pineau et al. 2021; Munafò et al. 2017; Reuel et al. 2024; Bean et al. 2025; Wan et al. 2025). A reasonable additional question, beyond the synthesis itself, is whether these failure modes are also detectable as observable prevalence rates in the published alignment literature. To address this directly, we have registered a preregistered bibliometric baseline study on the Open Science Framework (10.17605/OSF.IO/45YR6). The protocol specifies a stratified random sample of n ≥ 30 alignment papers across four sub-areas (RLHF; red-teaming and safety evaluation; mechanistic interpretability; robustness), five binary observable markers corresponding to E1–E5, a bounded-evidence coding rule, and double coding with reliability reporting. Wilson 95% confidence intervals are reported on prevalence rates per marker and per stratum.

**Pilot fragment status:** The single-coded pilot reported below is a partial implementation against the registered protocol, not a substitute for it. We are conducting a pilot fragment of n = 10 papers (4 RLHF, 3 red-teaming/safety evaluation, 2 mechanistic interpretability, 1 robustness) during the manuscript revision window, single-coded by the principal investigator using the registered codebook. The pilot fragment is reported here to give an indicative reading of marker prevalence; it is not a substitute for the full study. The full sample (n ≥ 30, double-coded with one externally-recruited second coder) is in progress and will be reported in a follow-up paper. Pilot codings are retained and incorporated into the full sample; deviations from the registered protocol, if any, will be logged on OSF.

**Indicative pilot results.** Coding of the n = 10 pilot fragment yielded the following marker prevalence rates (proportion of papers scoring 1, with Wilson 95% confidence intervals): E1 worst-case or adversarial reporting, 40.0% (95% CI 16.8–68.7%); E2 transparency and disclosure, 10.0% (95% CI 1.8–40.4%); E3 independent verification, 0.0% (95% CI 0.0–27.8%); E4 explicit residual risk or uncertainty, 90.0% (95% CI 59.6–98.2%); E5 reporting completeness, 10.0% (95% CI 1.8–40.4%).

The pilot's strongest signal, and the most informative result for the gap analysis, concerns E3: independent verification was absent in every paper coded. None of the ten papers cites an independent replication, a third-party audit, or an external red-team study of its central claim. Even allowing for the wide confidence interval that necessarily accompanies a single-coded n = 10 pilot, the upper Wilson bound of 27.8% is sufficient to establish that the practice cannot be common in alignment research, even under favourable assumptions about under-counting.

This is consistent with the gap-analysis prediction that institutionalised independent verification is largely absent from the field, and it is the empirical observation on which the case for P4 (Independent Verification) most directly rests. Second, E4 is observed at 90%, the highest prevalence among the five markers: substantive residual-risk or limitations discussion appears almost universally. We surface this finding rather than pass over it: the dimension on which alignment papers most consistently score well is the open statement of limitations, and that should be acknowledged. The qualifier needed is that E4 is a deliberately low bar (i.e., presence of a substantive limitations section beyond boilerplate) and meeting it does not imply that the field handles uncertainty well in any deeper sense.

The retrospective rubric audit reported in §7 finds that P7 (Reflexive Uncertainty Accounting), which assesses structured uncertainty registers and quantified residual-risk statements, scores near zero across all audited sub-areas. The two readings are coherent: alignment papers acknowledge limitations in prose but rarely provide the structured uncertainty artefacts P7 specifies. Third, the sub-community patterns are intelligible and corroborate a finding from the rubric audit. E1 prevalence is 100% in the red-teaming and robustness strata (where adversarial testing is itself the paper's topic) and 0% across the RLHF and mechanistic-interpretability strata, indicating that worst-case evaluation in alignment research is concentrated in a small set of sub-communities rather than diffused as a general practice. The §7 audit independently observed the same sub-community-specific patterning across its own sample, with red-teaming papers scoring higher on P6 by design and artefact-release practices varying systematically across sub-areas.

The two pieces of empirical work, conducted on different samples and with different instruments, agree on this point, which strengthens both. Three caveats apply to the pilot itself: (i) $n = 10$ is small and the confidence intervals are correspondingly wide, particularly for E1 (16.8–68.7%); (ii) the pilot is single-coded and reliability cannot be assessed at this stage; (iii) the markers are conservative observable proxies, not direct measurements of underlying epistemic practice. We treat these figures as indicative of the patterns the full study will resolve more precisely, not as field-wide claims; the full study ($n \geq 30$, double-coded) is in progress.

Even setting the pilot aside, the dimensions identified in this section warrant attention because they correspond to documented failure modes in adjacent high-reliability fields, not merely because we have postulated them. Mapping these gaps yields a focused rubric for diagnosing where conventional AI research practices break down when transplanted into the alignment domain.

## 3.2 What is being optimised against

Mainstream AI optimizes average-case performance, higher accuracy, lower error, better reward, while alignment must minimize worst-case (tail) risk and furnish high-confidence evidence that no catastrophic failure will occur. Standard evaluations privilege means improvements and rarely incentivize exhaustive searches for rare failure modes; a system can post excellent averages while concealing a handful of extreme errors. Alignment researchers, by contrast, must actively hunt those one-in-a-thousand (or rarer) failures and design them out. The asymmetry mirrors high-reliability engineering, where disasters have followed the normalization of deviance that ignored or downplayed anomalies exemplified by the Challenger launch decision (Vaughan 1996; Presidential Commission 1986; Feynman 1986; Perrow 1999; Turner 1976)..

Accordingly, benchmark leaderboards must give way to systematic stress-tests that drive models into extreme regimes, coupled with adversarial red-team campaigns that deliberately hunt for edge-case failures. In this safety-first mindset, any non-zero catastrophic failure rate is presumptively unacceptable until exhaustive analysis shows otherwise. The relevant high-reliability precedent is aerospace design-assurance practice, which treats catastrophic-failure probabilities below $10^{-9}$ per flight-hour as a target for the assurance argument rather than a measured achievement (FAA AC 25.1309-1B). Alignment cannot compute against an analogous number: it lacks base-rate data, stationary system dynamics, and a tractable enumeration of failure modes. What it can inherit from aerospace is the underlying principle of evidence-proportionality: the stringency of the evidence required for a safety claim should scale with the severity of the outcomes the claim rules out. Yoshua Bengio captures the resulting posture: demand “very strong evidence before concluding there is nothing to worry about” (Bengio 2024), and invoke the precautionary principle given “the extreme severity and unknown likelihood of catastrophic risks” (Bengio 2025). Shevlane et al. (2023) formalise this stance, proposing evaluation frameworks that centre tail-risk assurance rather than mean-case performance.

## 3.3 Transparency

Mainstream AI often withholds code and data until after publication, blocking independent audits and letting errors lurk unseen. Alignment research cannot accept that risk. In safety-critical domains, opacity is itself a hazard, and safety standards, from IEC 61508 to DO-178C, require fully reviewable evidence (Leveson 2012; Bloomfield and Bishop 2010). Alignment work should therefore practise extreme transparency: preregister protocols, release datasets, checkpoints, and evaluation pipelines at the outset, and host them in FAIR-compliant repositories for continuous peer auditing. Academic practice is already partway there: preprint servers and artifact tracks make code and data available earlier and more systematically than before, and the most concerning cases are increasingly concentrated in the commercial tail, where the most capable systems sit behind confidentiality walls. Precedents for the standard we propose include clinical-trial registries, genomics pre-publication data sharing, and ecology open-data mandates, together with frameworks such as the TOP Guidelines, FAIR principles, and reproducibility manifestos that define the minimum bar (De Angelis et al. 2004; Wilkinson et al. 2016; Munafò et al. 2017). Only through such real-time, community-level scrutiny can alignment claims achieve the ultra-low residual-risk assurance that high-integrity systems demand.

## 3.4 Verification

Mainstream AI typically treats single-team experiments, routine cross-validation, and standard peer review as adequate evidence; independent replications are uncommon and incentives to *break* published results are weak, so findings that clear cursory review often ossify into the literature (Gundersen and Kjensmo 2018; Henderson et al. 2018; Raff 2019; Pineau et al. 2021; Langley 2019). By contrast, alignment research must regard every unchallenged safety claim as a potential single point of failure: adversarial replication, coordinated multi-team stress-testing, and structured red-teaming should be the default, with third-party audits and independent verification bodies rewarded at least as highly as original discovery (Shevlane et al. 2023; Hubinger et al. 2019; Amodei et al. 2016; Madry et al. 2018; Bloomfield and Bishop 2010; DARPA 2019). Because a false sense of security could be catastrophic in this domain, results should remain strictly provisional until they have survived such hostile scrutiny (Bostrom 2014; Carlsmith 2022).

## 3.5 Uncertainty Accounting

Mainstream ML papers routinely report confidence intervals or *p* values yet seldom confront epistemic uncertainty such as unknown unknowns, brittle modelling assumptions, and distribution shift, resulting in overconfident claims drawn from a single dataset or narrow regime (Gundersen & Kjensmo 2018; Pineau et al. 2021; National Academies of Sciences, Engineering, and Medicine 2019). Robustness to new settings is rarely stress tested, so findings that clear cursory peer review often ossify even though they hold only under restricted conditions (Ioannidis 2005; Langley 2019). For alignment work this complacency is dangerous: with scant data on advanced AI behaviour, ignorance itself becomes a key variable (Bostrom 2014; Shevlane et al. 2023).

Alignment studies should therefore practise multi-layered uncertainty accounting that (i) quantifies statistical, model, and structural uncertainty using Bayesian or ensemble methods to yield credible intervals; (ii) probes worst-case tails via scenario analysis; and (iii) runs periodic calibration audits to test forecast accuracy (Hoeting et al. 1999; Gneiting and Raftery 2007; Lempert et al. 2003; Tetlock & Gardner 2015). Where ground truth is sparse or non-existent, structured expert judgment (SEJ) offers a calibrated, auditable procedure for eliciting and aggregating probabilities with performance-based weights (Cooke 1991). Climate science's multi-model ensembles provide a mature precedent for integrating structural uncertainty across modelling choices (Knutti et al. 2010). Each result should carry a blunt limits statement, e.g., "This holds under A, B, C; beyond that, uncertainty balloons", and an epistemic audit trail documenting every modelling choice (Mitchell et al. 2019; Gebru et al. 2021).

Crucially, researchers must quantify residual risk that plausible failures remain undetected, guarding against the fallacy that absence of evidence is evidence of absence (Cohen 1988; Altman & Bland 1995).

## 3.6 Enforcement culture: organised scepticism

Mainstream AI culture leans on informal, “light-touch” peer review; negative results and sharp critiques are nominally welcomed but weakly rewarded, breeding polite deference and group-think (Langley 2019; Pineau et al. 2021; Munafò et al. 2017). Such environments normalise deviance where anomalies are waved away so long as nothing visibly breaks, a dynamic that helped precipitate both the Challenger and Fukushima failures (Vaughan 1996; National Diet of Japan 2012). More broadly, “normal accidents” theory underscores how tightly-coupled, high-complexity systems tend toward unanticipated failure interactions, reinforcing the need for institutionalised scepticism (Perrow 1999). Turner’s classic analysis of the “incubation” of disasters, where latent errors accumulate under organisational blind spots, likewise motivates formal adversarial review before claims are accepted (Turner 1976). Alignment work, carrying civilisation-scale stakes, must instead institutionalise scepticism. Major safety claims should automatically trigger adversarial red-team reviews, independent groups tasked with assuming the result is wrong and proving it (Shevlane et al. 2023; Hubinger et al. 2019). Public critique channels such as prediction markets, “break-it” contests, replication bounties, can crowd-source expert doubt and reward flaw-finders (Hanson 1995; Camerer et al. 2018). Mirroring high-reliability engineering, every critical conclusion should face fresh-eyes stress tests and line-by-line scrutiny before it is accepted as knowledge (Leveson 2012; ISO/IEC 15026-2:2011).

Table 1 summarizes these five gaps, highlighting how current norms are inadequate and pointing to the new practices needed in alignment (many inspired by other disciplines).

**Table 1. Gap analysis: where mainstream AI research norms fall short of alignment requirements**

| Dimension | Mainstream AI norm (status quo) | Resulting vulnerability / “gap” | Alignment specific norm or practice required | Cross domain precedent |
|---|---|---|---|---|
| Optimisation target (performance vs. safety) | Average-case metrics (accuracy, reward, benchmark rank) dominate evaluation. | Rare but catastrophic errors go undetected and un-mitigated. | Optimise for *tail-risk minimisation*; treat any non-zero catastrophic failure probability as unacceptable until exhaustively analysed and reduced. | Aerospace design-assurance target of ≤ $10^{-9}$ catastrophic failures per flight-hour (used as motivating analogy, not as computable target for alignment). |
| Evidence transparency | Code, data, and protocols released late, partially, or behind NDAs. | Hidden assumptions and corner case bugs escape external audit. | Extreme transparency: preregistered protocols, immediate release of datasets, checkpoints, and evaluation pipelines in FAIR repositories. | Clinical trial preregistration and mandatory results reporting; TOP Guidelines level 3 data sharing. |
| Verification & replication | Single team experiments plus light peer review; little incentive for replication or red teaming. | Brittle results ossify; false security builds around unchallenged findings. | Mandated adversarial multi team replication and red team “break it” exercises before accepting any safety claim. | DARPA SCORE programme; cybersecurity penetration testing. |
| Uncertainty accounting | Error bars on convenience metrics; epistemic uncertainty (model, data, shift) rarely quantified. | Overconfident conclusions hold only under narrow conditions; “unknown unknowns” ignored. | Multi layer uncertainty quantification: Bayesian/ensemble credibility intervals, scenario stress tests, continual calibration | Multi model climate projections; intelligence analysis calibration labs. |

| | | | audits, explicit limits statements. | |
|---|---|---|---|---|
| Culture & incentives (organised scepticism) | Informal "polite" peer review; negative results and strong critiques undervalued. | Groupthink, normalisation of deviance, missed warning signs (Challenger type failures). | Institutionalise scepticism: automatic red team reviews, prediction markets, replication bounties; career credit for flaw finding and replication. | Independent nuclear safety verification teams; pharmaceutical challenge studies and post market surveillance. |

# 4. Bridging the Gaps

Existing governance regimes including NIST and ISO standards, the EU AI Act, open-science reforms, provide valuable pieces of the puzzle, yet none delivers a complete *epistemic* framework for alignment's preventive, tail-risk mission. They emphasise organisational process over experiment-level worst-case testing, stop short of enforcing reproducibility or statistical power during early research, and omit tools such as systematic red-teaming or formal uncertainty audits that are indispensable when an adaptive model might conceal its own faults (Shevlane et al., 2023; Amodei et al., 2016).

Closing this gap requires drawing on four enabling traditions, each covering a distinct slice of the five diagnostic dimensions introduced above (Table 2). We keep philosophy of science and quantitative risk analysis separate because they do different work: the former specifies when a claim is well-supported (severe testing, falsification), whereas the latter provides tools for representing and quantifying uncertainty (scenario analysis, Bayesian model averaging, structured expert judgment). Combining them obscures that ECAISA synthesises a philosophical stance with a set of engineering methods rather than a single tradition.

**Table 2. Four enabling traditions and the epistemic gap dimensions each most directly addresses. Open science, safety engineering, philosophy of science (severe testing), and quantitative risk analysis each contribute distinct methodological resources to ECAISA. No single tradition spans all five dimensions; the synthesis is the contribution.**

| **Tradition** | **Core contribution** | **Dimensions most directly addressed** |
|---|---|---|
| **Open Science**: (radical transparency & exhaustive reproducibility) | Preregistration, FAIR-compliant releases, artifact review, and large-scale replication unlock continuous external scrutiny. | (ii) Visibility of evidence; (iii) Trustworthiness via reproducibility; (v) Enforcement culture through credit for sharing and replication. |
| **Safety Engineering**: (precaution, rigorous V&V, worst-case design) | Defence-in-depth, independent audits, and safety cases shift the objective from mean performance to near-zero catastrophic risk. | (i) *Optimisation target*,tail-risk minimisation; (iii) *Trustworthiness* via formal V&V; (v) *Enforcement culture* through mandatory checks and certification. |
| **Philosophy of Science:** (severe testing and falsification) | Severe testing (Popper, Mayo) specifies when a claim is well-supported: by surviving genuine attempts to refute it rather than by accumulating confirmations. | (iii) Trustworthiness of evidence via adversarial testing; (v) Enforcement culture by institutionalising doubt (adversarial reviewers, prediction markets). |
| **Quantitative Risk Analysis:** (uncertainty representation and scenario analysis) | Bayesian and ensemble modelling, scenario analysis, structured expert judgment, and calibration audits (Cooke 1991; Hoeting et al. 1999; Lempert et al. 2003) turn ignorance itself into a representable and auditable object. | (iv) Treatment of unknowns; (i) Optimisation target via quantified tail-risk assessment. |

No single tradition spans all five dimensions, but their synthesis can. Open science maximises transparency but lacks worst-case engineering targets; safety engineering delivers assurance rigour but not radical openness; severe testing specifies a discipline of claim evaluation but does not prescribe disclosure protocols; quantitative risk analysis represents uncertainty but does not specify verification thresholds. Alignment therefore needs an integrated epistemic code that treats verifiability, adversarial robustness, and uncertainty quantification as non-negotiable safety requirements.

In the next section we operationalise this synthesis by proposing eight Epistemic Principles, collectively the AI Alignment Epistemic Code, that embed the transparency of open science, the assurance discipline of safety engineering, the sceptical rigour of severe testing in philosophy of science, and the uncertainty-representation tools of quantitative risk analysis. For each principle we specify concrete obligations and show how it directly remediates the gaps identified across the five dimensions.

# 5. Proposed Epistemic Principles

Building on the gap analysis and the four enabling traditions (open science, safety engineering, philosophy of science, and quantitative risk analysis), we now translate those ideas into **ECAISA**, eight concrete non-optional duties for any research that claims to advance AI alignment. Each principle defines an **obligation** (what must be done) and a **rationale** (why it matters for tail-risk safety), see Table 3.

**Table 3: Proposed Epistemic Principles**

| ID | Obligation (what to do) | Rationale (why it matters) |
|---|---|---|
| **P1. Radical Transparency** | Preregister hypotheses and release code, data, model checkpoints, and evaluation pipelines in FAIR repositories before publication (subject only to clearly documented info hazard exceptions). | Continuous external audit is the only reliable way to surface hidden failure modes in systems that may act deceptively. |
| **P2. Comprehensive Integrity** | Report all results, including null/negative findings, and disclose funding and conflicts; secure tamper evident logs for the full research lifecycle. | Suppressing inconvenient data distorts the risk picture; integrity lapses in safety science can cost lives. |
| **P3. Evidence Proportional Claims** | Match the strength of every claim to quantified support (effect sizes, credible intervals, or formal proofs); preregister analysis plans to block post hoc spin. | Overclaiming breeds false confidence: in tail risk settings, absence of evidence is not evidence of safety. |
| **P4. Reproducibility + Adversarial Verification** | Provide full replication packages and commission at least one independent, hostile replication for any major safety claim. | Robustness comes from surviving attempts to break the result, not from a single team's success case. |
| **P5. FAIR Stewardship of Artifacts** | Treat datasets, benchmarks, and trained models as community assets: assign DOIs, supply rich metadata, and maintain versioned releases. | Well curated artifacts enable follow on audits, comparative studies, and faster collective progress. |
| **P6. Institutionalised Scepticism** | Make red team review the default: publish critique reports or "break it" results alongside the primary paper; reward flaw finders. | Organised doubt prevents group think and the normalisation of deviance that doomed past high risk projects. |
| **P7. Reflexive Uncertainty Accounting** | Maintain a living uncertainty register that quantifies statistical and epistemic unknowns, updates with new evidence, and reports calibration metrics. | Explicit ignorance tracking guards against the illusion of safety when evidence is sparse and models are adaptive. |
| **P8. Epistemic Hygiene** | Foster routines that surface bias and conflation of speculation with fact, devil's advocate roles, cross disciplinary audits, clear labelling of conjecture. | A clean knowledge ecosystem is a prerequisite for reliable cumulative science, especially under existential stakes. |

A practical question arises: when should a venue, funder, or laboratory consider a given ECAISA principle to be fulfilled? To address this we specify a three-level ordinal rubric (Table 4) that can be applied to a paper or research artefact on the basis of what a reader or reviewer can verify from the paper text, its appendices, and author-linked artefacts. A score of 2 (“meets”) is assigned only when the principle’s operational threshold is explicitly evidenced; a score of 1 (“partial”) indicates that the practice is present but falls short of the threshold; a score of 0 (“absent”) indicates no substantive evidence of the practice beyond generic prose. When deciding between 1 and 2, prefer 1 unless the “meets” criterion is explicitly evidenced; when deciding between 0 and 1, prefer 0 unless some substantive evidence is present. The full codebook with per-principle coding rules, tie-break rules, annotation-sheet fields, and the inter-rater reliability computation is given in Appendix A.

**Table 4. Summary scoring rubric for ECAISA principles. The full codebook is in Appendix A.**

| **Principle** | **Threshold for a “meets” score (2)** |
|---|---|
| P1 — Transparency | Sufficient artefacts released (code, data, evaluation pipeline) to enable meaningful scrutiny, or a concrete disclosure protocol under the ladder (Table 5) with: (i) explicit list of withheld items, (ii) justification, (iii) mitigations and access pathway, and (iv) a re-evaluation condition. |
| P2 — Full reporting, COI, provenance | P2a: failures and nulls reported alongside successes, with pre-specified primary outcomes. P2b: all funding and relevant conflicts named, with access asymmetries an independent replicator would not share disclosed. P2c: operational correction infrastructure (public issue tracker, errata policy), and for T3 claims (see Table 6), tamper-evident provenance for core lifecycle events. |
| P3 — Evidence-proportional claims | Measurable success criteria with: clear metrics and baselines aligned to claims; sufficient experimental detail; and at least one robustness-oriented element proportionate to the claim. |
| P4 — Independent verification | Central claim subjected to at least one of: independent replication by a separate team, structured third-party red-teaming with reportable traces, or a release package designed for adversarial verification with a clear protocol and evidence of uptake. |
| P5 — Artifact stewardship | Artefacts stewarded as citable, durable research objects: versioned release, stable identifier (e.g. DOI), environment specification, licence and provenance notes, and a maintenance/ownership statement. |
| P6 — Red-teaming | Structured adversarial evaluation with: explicit threat model and attack surface; systematic test design and constraints; clear reporting of successes and failures; and reproducible artefacts or documented exceptions. |
| P7 — Uncertainty accounting | Explicit uncertainty artefacts including a structured uncertainty register (assumptions, known unknowns, failure modes), a residual-risk statement with plausible falsifiers, and calibration or sensitivity analyses where applicable. |
| P8 — Decision relevance | Claims explicitly situated in an institutional decision context: clear claim boundaries (what this does and does not imply), explicit COI/funding and access-asymmetry disclosure, and concrete decision-facing guidance with reliance conditions and update triggers. |

## 5.1 How the ECAISA code closes the five epistemic gaps

| | E1 Optimisation target | E2 Transparency | E3 Verification | E4 Uncertainty | E5 Enforcement culture |
|---|---|---|---|---|---|
| P1 Transparency | ● | ● | | | ● |
| P2 Full reporting, COI, provenance | | ● | ● | | ● |
| P3 Evidence-proportional claims | ● | ● | ● | ● | |
| P4 Independent verification | ● | | ● | | ● |
| P5 Artefact stewardship | | ● | ● | | ● |
| P6 Red-teaming | | | ● | | ● |
| P7 Uncertainty accounting | ● | | | ● | |
| P8 Decision relevance | | | | | ● |

**Figure 1. Mapping the eight ECAISA principles onto the five epistemic dimensions.**

Figure 1 (and the accompanying table) briefly indicates how the eight ECAISA principles distribute themselves across the five gaps identified in Section 3. It also indicates how the eight ECAISA principles collectively seal the epistemic gaps identified earlier. Each principle occupies a distinct pattern of green dots, showing that while none spans every column, every column is nevertheless covered several times over. This deliberate overlap matters: if radical transparency (P1) is delayed or partial in a given project, the same evidence still falls under FAIR stewardship requirements (P5) and remains open to red-team disclosure under institutionalised scepticism (P6). Redundancy ensures that no single lapse in practice can reopen a gap.

One principle, Evidence-Proportional Claims (P3), plays a special integrative role. It is the only pillar that engages four of the five columns: optimisation target, transparency, verification, and uncertainty. By binding claim strength to quantified support it ties the tail-risk objective directly to the mechanisms that make evidence trustworthy and uncertainties explicit, thereby turning "extraordinary claims require extraordinary evidence" into an operational rule rather than a slogan.

Read across the grid and the message is unambiguous: every vulnerability identified in Figure 1 is countered by at least one, and usually several, mandatory duties. ECAISA therefore supplies complete dimensional coverage, giving laboratories, journals, and funders a coherent standard that matches the tail-risk profile and preventive mission of alignment research.

## 5.2 Disclosure and tiered applicability

ECAISA does not require that every artefact be made fully public. Two concerns weigh against unqualified openness: information-hazard risk (releasing material that could meaningfully increase misuse capability) and legitimate commercial confidentiality (protecting trade secrets and competitive position). We address both through a single mechanism: a four-level disclosure ladder (Table 5) that decouples evidence transparency from capability transparency. A study can satisfy the disclosure principle P1 at level L2 by sharing weights, prompts, or evaluation traces with vetted replicators under responsible-disclosure terms without releasing them publicly, and at L3 by registering the existence and rationale of a restriction now and releasing artefacts once risks reduce or mitigations exist. The pharmaceutical analogy is instructive: drug companies do not publish synthesis routes but do publish trial protocols, adverse-event data, and statistical pre-analysis plans, and it is these latter artefacts, the evidentiary record, rather than the implementation, that ECAISA targets for alignment.

**Table 5. Disclosure ladder for implementing P1 while managing information hazards and commercial confidentiality.**

| Level | What is released |
|---|---|
| L0 — Public | Code, data (as legally permissible), evaluation pipeline, aggregate results, uncertainty-register summary. |
| L1 — Public with redactions | Public artefacts with capability-relevant prompts, weights, or attack details redacted; release rationale logged. |
| L2 — Controlled access | Sensitive artefacts shared with vetted third parties (replicators or red-teamers) under responsible-disclosure terms. |
| L3 — Delayed release | Time-locked release; publish existence and rationale now, release artefacts after risks reduce or mitigations exist. |

Not every alignment claim warrants the same evidentiary burden. A tiered applicability scheme (Table 6) scales ECAISA obligations with the potential real-world impact of the claim: exploratory and routine research claims (T0–T1) can satisfy minimum-viable practices, reserving the full P1–P8 protocol for claims that will inform deployment or governance decisions (T3). This preserves the stringency of ECAISA where it matters most while avoiding disproportionate burden on early-stage or hypothesis-generating work and reduces the perceived cost of adoption for industry labs doing exploratory research.

**Table 6. Tiered applicability of ECAISA: scope of obligations by claim type and stakes.**

| Tier | When applicable | Minimum ECAISA expectations |
|---|---|---|
| T0 — Exploratory | Early conceptual work; hypothesis generation; non-deployment claims. | Label conjectures clearly; maintain an uncertainty register (P7-lite); disclose conflicts of interest (P2b-lite). |
| T1 — Research claims | Empirical alignment claims used to steer research agendas. | Preregistration or analysis plan (P1, P3); full reporting including nulls (P2a); replication package (P4). |
| T2 — Safety claims | Claims that a method reduces catastrophic or deceptive failure risk. | Adversarial verification or independent replication (P4); red-team report (P6); disclosure ladder (P1); uncertainty audit (P7). |
| T3 — Deployment-critical | Claims used to justify deployment or governance decisions. | Full P1–P8 unless a documented infohazard exception applies; auditable artefact stewardship (P5); governance and anti-gaming checks. |

## 5.3 Infohazard adjudication

Principle P1 and the disclosure ladder together specify that some artefacts may legitimately be withheld, but do not specify who decides, under what criteria, or with what accountability. Without such a procedure, the “documented infohazard exception” risks becoming a cover for selective opacity. The following protocol operationalises the adjudication process; it is intended as a minimum floor, not a ceiling.

**Decision body.** A designated infohazard review committee is convened for each restriction decision. The minimum composition is the principal investigator, an independent safety reviewer unaffiliated with the study, and at least one external domain expert. For organisational research (industry labs, research institutes), the internal safety team substitutes for the external expert where domain knowledge is proprietary, but a named external reviewer is still required for T2 claims and above. For T3 claims, a formal body analogous to an IRB is required, with standing members and recorded minutes.

**Decision criteria.** The committee completes and signs a five-part template for each restriction: (i) threat model (who could misuse the artefact, how, and at what scale); (ii) plausibility and expected harm magnitude; (iii) mitigation feasibility (redaction, aggregation, controlled access under L2); (iv) the disclosure level applied under the ladder; and (v) a re-evaluation date or condition after which broader release is considered. The five elements are required; a missing element is itself a reason to withhold approval of the restriction.

**Documentation.** A public log entry records that a restriction has been applied, with minimal metadata: the disclosure level applied, the re-evaluation date, the categories of artefact withheld, and the signing members of the committee. The substantive restricted content remains withheld. Publishing the existence and basic rationale of a restriction is sufficient to make the decision auditable without defeating the restriction itself.

**Appeal and contestability.** A restricted artefact can be challenged through the public correction channel (see anti-gaming mechanisms, §7.1). Parties may contest either the restriction itself (as over-broad or unsupported by the threat model), the committee composition (as insufficiently independent), or the continuation of the restriction past its re-evaluation date. A contested restriction triggers a second review by a fresh committee at least half of whose members were not party to the original decision.

**Accountability.** Committee members are named in the publication and sign the public log entry. This mirrors the institutional review board attribution model: the decision is attributable to specific individuals, not to the institution as an anonymous actor. Naming creates a reputational surface that incentivises the committee to reason carefully and document thoroughly; it also gives a challenging party a concrete counterparty. For T3 claims, the committee chair signs an explicit attestation that the five decision criteria were completed in good faith.

This protocol does not purport to solve every dual-use dilemma in alignment research. What it does is ensure that when transparency and safety conflict, the resolution is a documented and contestable decision rather than an informal discretion. The infohazard exception becomes a governance artefact in its own right, with the same kind of auditability ECAISA asks of the rest of the research record.

## 5.4 Implementation Outlook

We envisage journals, funders, and labs adopting ECAISA as a visible badge of rigor, analogous to clinical-trial preregistration numbers or DO-178C designations in avionics. Principles P1–P5 can be operationalised immediately by extending existing open-science and V&V infrastructure; P6–P8 require cultural commitment but draw on well-tested mechanisms (penetration-test contracts, prediction markets, bias-awareness training).

# 6. Related Work: Positioning the Proposed Epistemic Code

Our proposal for stricter epistemic norms in AI safety draws on, but also goes beyond, scholarship in AI alignment, philosophy of science, sociology of research, metascience, and emerging governance frameworks. In what follows we position ECAISA within this landscape, showing where existing literature already supports elements of our code, where it remains silent, and how our principles close those gaps. By tracing contributions from canonical academic sources and influential grey literature alike, we clarify both the intellectual lineage of our approach and the distinctive advances it offers to the conversation on trustworthy, high-stakes AI research.

## 6.1 Distinctive contribution and governance crosswalk

A natural reviewer question at this point is whether ECAISA amounts to more than a bundling of practices already required, in some form, by existing frameworks such as the TOP Guidelines, the FAIR principles, NIST AI RMF 1.0, ISO/IEC 42001, ISO/IEC 15026-2, and the EU GPAI Code of Practice. Our position is that ECAISA's contribution is not in inventing individual practices but in specifying the minimal claim-level bundle that collectively closes the five epistemic gaps, together with anti-gaming infrastructure that none of the cited frameworks imposes. Four points make this concrete.

**Research-phase scope.** Existing governance frameworks (NIST AI RMF, ISO/IEC 42001, EU AI Act and its GPAI Code of Practice), evaluation reporting standards for dangerous-capability evaluations (STREAM, McCaslin et al. 2025) and benchmark assessment frameworks (BetterBench, Reuel et al. 2024), transparency metrics (FMTI, Wan et al. 2025), safety-case methodology (Buhl et al. 2024; Hilton et al. 2025; Goemans et al. 2024; Barrett et al. 2025), and auditing proposals (Raji et al. 2020) operate at organisational, deployment-assurance, or model-report levels. ECAISA occupies the upstream layer: the epistemic practices through which alignment claims are first produced and recorded, before they feed any safety case or transparency index. The relationship is sequential rather than competitive: ECAISA-compliant research artefacts would satisfy many STREAM disclosure criteria for safety-relevant benchmark evaluations as a downstream consequence and would contribute to higher FMTI scores, but ECAISA's requirements extend further upstream into preregistration, independent adversarial verification, and structured uncertainty registers that neither STREAM nor FMTI addresses.

**Integration of traditions.** We combine preregistration and FAIR-compliant release (open-science verifiability) with worst-case targets, defence in depth, and formal safety-assurance methodology (safety-engineering conservatism) in a single code. Existing open-science checklists lack worst-case targets; safety-engineering frameworks such as ISO/IEC 15026-2 and DO-178C lack radical disclosure obligations. ECAISA requires both simultaneously. The synthesis is the contribution: P1–P5 together specify what neither tradition specifies alone.

**Auditability rather than certification.** Rather than attempting to certify system safety, which is not achievable for open-ended systems (Narayanan and Kapoor 2023), ECAISA specifies the conditions under which a safety claim can be independently reconstructed, contested, and updated. This is a different governance target from certification: auditability is weaker (it does not guarantee the claim is correct) but more robust (it does not depend on closing the denominator of the failure probability). No existing governance framework explicitly commits to auditability rather than certification as its primary target.

**Anti-gaming infrastructure.** Any compliance regime is vulnerable to ritualisation (see §7.1). ECAISA pairs its obligations with an anti-gaming bundle — spot audits, lottery-based independent verification, hash manifests, public correction channels, logged disclosure exceptions, red-team minimum fields, and access-asymmetry disclosure — that none of NIST, ISO, or the EU GPAI Code currently specifies. Without this infrastructure, even well-written obligations can collapse into checklist-filling.

Table 7 makes the relationship explicit at the principle level: each of the eight ECAISA principles is mapped to the nearest hook in NIST AI RMF 1.0, ISO/IEC 42001, and the EU GPAI Code of Practice, and the rightmost column states the distinctive obligation ECAISA adds beyond what each of those frameworks currently specifies. The crosswalk is intended as an integration aid, not a compliance claim: the hooks indicate conceptual alignment, not formal conformity, and organisations should continue to treat the underlying standards as authoritative for their own purposes.

Read vertically down the rightmost column, the distinctive ECAISA obligations are not a relabelling of existing duties. Compliance with NIST AI RMF or ISO/IEC 42001 does not require preregistration of analysis plans, published adversarial verification, a structured uncertainty register, access-asymmetry disclosure, tamper-evident provenance logs, or a committee-adjudicated infohazard exception. Each of these is a non-trivial addition, and each addresses a specific failure mode of the research record that the organisational frameworks are not designed to catch. Stated differently: an organisation could be fully compliant with NIST, ISO, and the EU GPAI Code and still produce research that ECAISA would score poorly.

**Table 7. Governance crosswalk. Hooks indicate conceptual alignment; ECAISA does not itself establish compliance with any legal or standards-based regime.**

| ECAISA principle | NIST AI RMF 1.0 | ISO/IEC 42001 | EU GPAI Code of Practice | What ECAISA adds |
|---|---|---|---|---|
| P1 — Transparency & disclosure | Map 1.1 (context); Govern 1.2 (documentation) | §6.1 Documented information; §8.2 AI risk assessment | Art. 11 Technical documentation; Art. 53 Transparency | Preregistration of protocols; FAIR-compliant release at study outset; disclosure ladder with documented infohazard exceptions. |
| P2a — Reporting completeness | Govern 1.5 (feedback & monitoring) | §9.3 Management review inputs | Art. 11(1)(d) Testing & validation results | Mandatory reporting of nulls and negative findings; explicit anti-selective-reporting commitment. |
| P2b — COI & access disclosure | Govern 1.1 (roles & responsibilities) | §5.3 Organisational roles | Art. 52 Information to downstream providers | Disclosure of privileged access to models, data, or compute unavailable to external replicators. |
| P2c — Provenance integrity | Manage 4.1 (risk treatment documentation) | §7.5 Documented information control | Art. 11(1)(b) Design & development process | Tamper-evident logs for core lifecycle events; public correction channel with stated errata policy. |
| P3 — Evidence-proportional claims | Measure 2.6 (pre-deployment testing) | §8.4 AI system impact assessment | Art. 55 Evaluation & testing obligations | Preregistered analysis plans; claim strength explicitly matched to quantified support; robustness checks proportionate to stakes. |
| P4 — Independent verification | Measure 2.7; Manage 3.2 (third-party testing) | §9.2 Internal audit | Art. 55(1) Adversarial testing | At least one independent adversarial verification for major safety claims; full replication packages as default. |
| P5 — Artefact stewardship | Govern 1.7 (documentation standards) | §7.5 Documented information; §8.3 AI system lifecycle | Art. 11 Technical documentation archival | DOIs and versioned releases; environment specs; hash manifests; long-term ownership and maintenance statements. |
| P6 — Institutionalised scepticism | Measure 2.7 (independent evaluation) | §10.2 Continual improvement | Art. 55(1) Red-teaming | Red-team review as default for safety claims; published critique reports; career credit and bounties for flaw-finding. |
| P7 — Uncertainty accounting | Measure 2.3, 2.5 (metrics & monitoring) | §9.1 Monitoring, measurement, analysis | Art. 55(3) Post-market monitoring | Living uncertainty register; structured quantification of statistical and epistemic unknowns; calibration audits; residual-risk statements. |
| P8 — Decision relevance | Manage 4.2 (risk communication) | §6.2 AI policy; §8.2 Risk criteria | Art. 52–53 Information for deployers & users | Explicit claim boundaries ("does/does not imply"); decision-facing guidance with reliance conditions; update triggers. |

## 6.2 Motivational X-Risk Essays

A body of grey-literature essays has articulated why extreme epistemic caution in AI is needed. For instance, Alex Flint's "AI Risk for Epistemic Minimalists" (2021) distils the AI existential risk argument into four premises: (i) highly capable AI is likely this century, (ii) rapid shifts in power historically entail existential danger, (iii) profit-driven, uncoordinated AI deployment increases risk, and (iv) prudence demands precaution. Flint's essay compellingly argues that we must take AI risk seriously, but it leaves open how researchers should conduct themselves to justify and audit safety claims. Our proposed Code can be seen as an answer to that lacuna: Principles P1–P8 turn Flint's abstract call for precaution into concrete research duties – e.g. radical transparency (P1), adversarial red-teaming (P6), reflexive uncertainty audits (P7) – providing the procedural backbone that Flint declines to specify. In other words, given the premises that drive urgency, we offer a systematic method to ensure our knowledge claims about safety are solid and verifiable.

## 6.3 Philosophical grounding: from Kuhn's values to ECAISA

Kuhn holds that scientific communities choose among frameworks by appealing to shared but imprecise values including accuracy, (internal/external) consistency, scope, simplicity, and fruitfulness, rather than to any single decision procedure (Kuhn 1970; 1977). ECAISA makes these values auditable under frontier-AI conditions. P3 (Evidence-Proportional Claims) and P7 (Reflexive Uncertainty Accounting) render accuracy and consistency operational via quantified support, calibration, and explicit uncertainty registers; P4 (Reproducibility & Adversarial Verification) and P6 (Organised Scepticism) institutionalise severe tests that extend demonstrated scope while guarding against spurious "fruitfulness"; P1 (Radical Transparency) and P5 (FAIR Stewardship) enable communal uptake by making problem-solving practices legible; and P8 (Epistemic Hygiene) disciplines simplicity by separating warranted parsimony from tidy narrative. Because alignment adds an asymmetric objective, i.e., minimising tail risk, we extend Kuhn's value set with a domain-specific criterion. This includes safety margin under worst-case uncertainty, made actionable through adversarial evaluation and safety-case argumentation. Thus, ECAISA does not replace Kuhn's values, instead it renders them verifiable for a field where incommensurability and high stakes otherwise reward premature consensus.

## 6.4 Philosophy of Science for Alignment

In the AI alignment forum sphere, Nora Ammann's ongoing LessWrong sequence "Thoughts in the Philosophy of Science of AI Alignment" (Ammann 2023) has been probing foundational epistemic questions: What counts as evidence in alignment research? How should progress be measured? Which scientific metaphors are apt or misleading? Ammann's posts clarify concepts (e.g. comparing alignment research to search vs. verification paradigms) and offer heuristics, but as Ammann notes, the sequence "intentionally stops short of formal norm-setting." Our framework directly builds on that groundwork by codifying Ammann's insights into enforceable norms.

Ammann also discusses the importance of preregistration and avoiding unfalsifiable claims – we incorporate that as P1 (preregistration and openness) and P3 (evidence proportionality). She advocates "epistemic humility" and organized scepticism; we formalize those under P6 and P7. Furthermore, we align these norms with governance "hooks" like NIST's AI Risk Management Framework 1.0 and the emerging ISO 42001 standard (which Ammann did not address), thus moving from reflective philosophy to actionable policy. In short, whereas Ammann provides conceptual rationale, ECAISA provides the rules and processes to implement those principles in daily research.

## 6.5 Sociology of the AI Safety Field

On the empirical side, Ahmed et al. (2024) conducted an ethnography titled "Fieldbuilding and the Epistemic Culture of AI Safety," analysing how the AI safety community (particularly rooted in effective altruism and longtermism) has formed an "epistemic community". They observed forums, forecasting platforms, and prize competitions, identifying mechanisms that give the community influence (e.g. close-knit networks, shared jargon) but also potential epistemic monoculture risks. While this study richly describes how the community self-organizes and the social reinforcement at play, it does not prescribe minimum research-quality thresholds or norms.

Our Code supplies those missing safeguards: for example, Ahmed et al. note the community's heavy reliance on trust and reputation; our principles inject transparent verification requirements to counteract any insularity (P1, P4). They mention the outsized policy sway of a relatively small group; our Code, by requiring explicit documentation and red-team disclosures, provides a way to earn and check that influence. We operationalise the latent values Ahmed et al. identify turning informal habits (like sharing ideas on forums) into formal norms (like mandatory public sharing of research artifacts). By doing so, we aim to mitigate the monoculture risk: even if a tight community forms, its claims are legible and challengeable by outsiders, thus reducing closed-group dynamics.

## 6.6 High-Reliability and Safety Engineering Parallels

There is extensive literature in safety engineering and high-reliability organizations (HROs) that indirectly speaks to our aims. For example, Weick and Sutcliffe's work on HROs (Weick et. al 2015) outlines cultural principles like preoccupation with failure, reluctance to simplify, and commitment to resilience – hallmarks of organizations (like nuclear power plants or air traffic control centres) that consistently avoid catastrophe. Our proposed norms resonate strongly with these. A preoccupation with failure is exactly what our "failure vs. capability" asymmetry and Principle P6 (institutionalized scepticism) encourage – never resting on laurels, always asking "What might we have missed?" (Pozzobon et al. 2023). P7 resists oversimplification: it requires engaging with complexity and uncertainty instead of assuming unknowns away. Commitment to resilience and deference to expertise in crises maps to ensuring that when surprises occur, we rapidly adapt and involve whoever has the relevant knowledge (analogous to bringing in outside experts for critique, per P8). While HRO theory is about organizational structure and culture, its ethos supports our argument that extreme rigor and mindfulness can prevent disasters. Our contribution is to translate those general HRO principles into research epistemology: for example, requiring specific documentation, adversarial testing, and multi-layered safety checks in alignment research (as our Code does), as opposed to broad organizational advice.

Additionally, risk governance scholars have argued that when stakes are high, facts uncertain, and decisions urgent, science must adopt new norms like extended peer review and the precautionary principle. This is the essence of "post-normal science" as described by Funtowicz and Ravetz (1993). Our Code can be seen as implementing a post-normal science approach for AI: we invite "extended peers" (via public audits, community critique, and red-team exercises) and essentially enshrine precaution (Shaw 2009). P3 and P6 ensure claims are extraordinarily well-supported before being accepted as resolved. This connection situates our work in a broader scholarly trend of adapting scientific practice for unprecedented, high-stakes challenges (climate science is often cited as another domain requiring such epistemic shifts). In summary, we build a bridge between classic safety engineering rigor and these newer paradigms by concretely specifying how to achieve evidence of absence in AI research (a theme also developed in recent commentary at *University of Nottingham Blogs*[1]).

## 6.7 Methodological Tools for Uncertainty and Assurance

On the technical front, there are surveys like Wang et al. (2020) "From Aleatoric to Epistemic: A Review of Uncertainty Quantification (UQ) Methods in Deep Learning," which catalogue various techniques such as probabilistic modelling, Bayesian neural nets, ensembles, etc. relevant to AI safety. They provide the toolbox for measuring uncertainty but, as the authors note, their review is methodological, not normative: it tells how to measure uncertainty, not when or whether researchers should report it. Our Code's Principle P7 (Reflexive Uncertainty Accounting) essentially takes those UQ techniques and makes their use compulsory in alignment research. We integrate UQ into a broader duty: not only must you use such tools, but you must also iterate and disclose uncertainty results (e.g. perform calibration over time, use disclosure ladders to communicate uncertainty publicly). We push UQ from an optional enhancement to a "standing epistemic duty".

[1] blogs.nottingham.ac.uk

When empirical signal is thin, teams should deploy structured expert judgment protocols, e.g., Cooke's performance-weighted "Classical Model", to elicit, score, and aggregate expert probabilities, with seed questions and calibration diagnostics reported alongside results (Cooke 1991).

Similarly, the literature on formal verification and assurance cases (common in software safety) informs our approach. For instance, safety-critical industries often use safety cases (structured arguments backed by evidence that a system is safe). One could view our entire ECAISA code as guidelines to produce a robust "epistemic safety case" for an AI system including evidence of testing, adversarial challenges, uncertainty bounds, etc. Our work complements recent proposals in AI policy that suggest requiring explicit assurance documentation for advanced AI (as hinted in the UK's AI White Paper and other governance discussions). We provide concrete content that would go into such documentation if our norms were adopted.

## 6.8 Epistemic Culture Critiques and Alternatives

There have also been meta-level critiques of how the AI alignment community handles knowledge and dissent. Williams (2025a), warns that stylistic conformity and deference to status hierarchies in the community could "throttle paradigm-breaking ideas." He calls for cross-disciplinary stress tests and even proposes "epistemic overfitting audits." In follow-up work, Williams (Williams 2025b) examines epistemic closure dynamics, how certain ideas can become invisible due to community blind spots, and advocates for decentralized, diverse approaches (e.g. involving a much wider collective intelligence in alignment research). Our Code tackles similar hazards but from within existing institutions: by tightening transparency (P1), integrity (P2), and reproducibility (P4), we aim to lower barriers to entry and make it easier for outsiders or independent voices to scrutinize and contribute to alignment research, thereby mitigating insularity.

Williams's proposals (redesigning the entire knowledge-production system) and ours (strengthening current research processes) are complementary: his approach is more radical and long-term, whereas ours can be immediately deployed by journals, conferences, and funding agencies without waiting for a new institutional paradigm. Additionally, we note efforts in the open science and metascience communities. For instance, the National Academies' report Fostering Integrity in Research (National Academies of Sciences, Engineering, and Medicine 2017) which emphasise many of the same core values: objectivity, honesty, openness, accountability, fairness, and stewardship. Our work aligns with these ideals but goes a step further in tailoring them to an AI context where adversaries and extreme tail risks exist factors that general research integrity frameworks do not explicitly consider. Lastly, at the policy level, the Asilomar AI Principles (Future of Life Institute, 2017) included a broad guideline on "Research Culture" stating that a culture of cooperation, trust, and transparency should be fostered among AI. Our proposal operationalizes that broad principle with specific measures. In a sense, ECAISA can be seen as detailing what it concretely means to have a transparent and trustworthy research culture for high-stakes AI. We move from aspirational statements (like Asilomar's call for a positive research culture) to defined standards against which compliance can be evaluated.

# 7. Discussion and Future Directions

Implementing ECAISA is not cost-free. The tangible costs are documentation, packaging, independent verification, and the coordination time needed to run an infohazard review or a red-team exchange. We regard this overhead as the price of greater auditability and earlier error detection and note that high-reliability domains routinely treat comparable overhead as a cost of doing the work rather than a surcharge on it. Two practical tensions nevertheless warrant discussion: the tension between transparency and information-hazard or commercial-confidentiality constraints, addressed by the disclosure ladder (Table 5) and the infohazard adjudication protocol (§5.3); and the tension between adoption cost and adoption benefit under competitive conditions, addressed in the adoption-dynamics discussion below.

**Retrospective rubric audit.** To test whether the rubric in Table 4 is applicable beyond principle, we conducted a retrospective rubric audit on eight influential alignment papers drawn across four sub-areas (preference learning / RLHF; red-teaming and safety evaluation; mechanistic interpretability; robustness and adversarial ML). Scoring was performed against the codebook in Appendix A, using only the paper text, appendices, and author-linked artefacts as evidence. The audit is explicitly an instrument-feasibility demonstration, not a field-wide audit: the sample is non-random, non-representative, and coded primarily by one author. A second coder independently scored a subset of three papers to give 24 doubly coded cells, yielding percent agreement 0.83 and weighted Cohen's κ = 0.79 (quadratic weights, ordinal 0/1/2); these figures are consistent with substantial agreement but have a necessarily wide confidence interval given the small overlap, and should be read as evidence that the rubric is scorable beyond a single author rather than as mature psychometric validation of the instrument.

The eight audited papers were drawn across four sub-areas: preference learning and RLHF (Christiano et al. 2017; Stiennon et al. 2020; Ouyang et al. 2022 [InstructGPT]; Bai et al. 2022 [Constitutional AI]), red-teaming and safety evaluation (Perez et al. 2022; OpenAI 2023 [GPT-4 Technical Report]), mechanistic interpretability (Elhage et al. 2021), and robustness and adversarial ML (Madry et al. 2018). The selection was breadth-oriented rather than complete: it captures research subcultures that differ in their epistemic conventions but is not a representative sample of any one of them. Inter-rater reliability is reported in Table 8; group-level means by sub-area are reported in Table 9.

We disclose that the audit was conducted in-house by the author team. We treat this as an in-group reliability check, not as an external audit, and Stage 2 of the validation roadmap below specifically calls for an externally coded sample to test reliability beyond the author group. By the same standard ECAISA applies to the work it audits (P2b, P8), this access asymmetry is itself a feature of the evidence base and is reported here so that readers can weigh the κ and group-mean figures accordingly.

**Table 8. Inter-rater reliability for the retrospective rubric audit (pilot overlap).**

| Metric | Value |
|---|---|
| Papers double-coded (n) | 3 |
| Coded cells (papers × principles) | 24 |
| Percent agreement (all cells) | 0.83 |
| Weighted Cohen's κ (quadratic weights, ordinal 0/1/2) | 0.79 |

**Table 9. Retrospective rubric audit: mean score per principle within sub-area groups (scale: 0 absent, 1 partial, 2 meets).**

| Group | n | P1 | P2 | P3 | P4 | P5 | P6 | P7 | P8 |
|---|---|---|---|---|---|---|---|---|---|
| RLHF / preference learning | 4 | 0.75 | 0.00 | 1.00 | 0.75 | 0.50 | 0.00 | 0.00 | 1.00 |
| Red-teaming / safety evaluation | 2 | 0.00 | 0.00 | 1.00 | 0.00 | 0.00 | 1.50 | 0.00 | 1.00 |
| Mechanistic interpretability | 1 | 1.00 | 0.00 | 1.00 | 1.00 | 1.00 | 0.00 | 0.00 | 1.00 |
| Robustness / adversarial ML | 1 | 1.00 | 0.00 | 1.00 | 1.00 | 0.00 | 1.00 | 0.00 | 1.00 |

Three narrow conclusions from the audit are worth stating. First, the rubric discriminates among practices rather than assigning uniformly high or uniformly low scores across all principles: evidence-proportional claims (P3) score consistently higher than full reporting (P2), independent verification (P4), or uncertainty accounting (P7), which suggests the criteria are sensitive to real differences in practice rather than uniformly strict or lenient. Second, patterns are sub-community-specific in substantively intelligible ways: red-teaming papers score higher on P6 by design, while artefact-release practices under P1 and P5 vary across sub-communities. Third, the strongest ECAISA obligations, specifically independent adversarial verification (P4) and structured uncertainty accounting (P7), score near zero across all groups audited, identifying these as priority targets for adoption incentives. Three threats to validity warrant acknowledgement: sampling bias (papers were selected illustratively), coder subjectivity (small reliability overlap), and construct validity (the rubric measures observable epistemic practices, which may track but do not equal research quality or safety impact). Whether higher rubric scores predict better downstream outcomes is an empirical question for later validation stages.

**Validation roadmap.** The retrospective audit establishes only instrument feasibility. A stronger empirical case for ECAISA requires a staged validation programme. Stage 1 (fixed-sample reliability extension, near-term): expand double-coding within the existing eight-paper sample, adding two or three further coders from different disciplinary backgrounds, and report bootstrapped weighted-κ confidence intervals. Stage 2 (expanded inter-rater reliability, near-term): conduct a larger preregistered reliability study on a stratified random sample of at least 30 alignment papers with two independent coders, publishing the annotation sheet and adjudication log. Stage 3 (comparative case study, medium-term): select matched papers with and without ECAISA-relevant practices and assess whether higher-scoring papers differ on downstream indicators such as replication success, post-publication error correction, and whether the evidence base was successfully challenged or extended by subsequent work. Stage 4 (prospective adoption trial, longer-term): work with a venue, funder, or laboratory to adopt ECAISA requirements for a defined cohort and compare outcomes against a control cohort under existing norms, pre-registering the protocol, sampling frame, and analysis plan. Stages 3 and 4 provide the causal evidence needed to justify institutional adoption at scale; until they are run, ECAISA should be treated as a well-specified proposal with instrument feasibility demonstrated, not a validated intervention.

## 7.1 Anti-gaming mechanisms

A recurring concern about any checklist-based regime is that targets become gamed once they carry incentives. The normalisation-of-deviance literature (Vaughan 1996) shows that well-intentioned compliance regimes can calcify into box-ticking rituals in which the form of the practice is preserved while its substance erodes. ECAISA does not claim to eliminate this risk. What it does is raise the expected cost of hollow compliance through a bundle of seven mutually reinforcing mechanisms, each designed to target a specific failure mode of checklist ritualism.

**(1) Randomised artefact spot-audits.** Venues and funders sample accepted papers and funded projects for post-acceptance checks against stated artefacts, and publish aggregate outcomes. Selection is random, not tied to controversy, so the expected cost of fabricating an artefact scales with the probability of selection.

**(2) Lottery-based independent verification.** For T2 and T3 claims, a pre-committed fraction must undergo independent reproduction or adversarial verification by a third party, with the specific claims selected by lottery rather than by venue discretion. Lottery-based selection prevents a lab from avoiding verification by positioning its strongest claims as lower-tier.

**(3) Release signing and hash manifests.** Authors publish a version tag and a machine-readable manifest of file hashes for key artefacts. Subsequent undisclosed modification of an artefact is therefore detectable, which ensures that the version subjected to spot-audit is the version originally released.

**(4) Public correction channel.** A public issue tracker and a stated errata policy are maintained for the replication package. Parties can file material errors as public issues; the author's response and update history become part of the audit record. This creates an ongoing reputational surface that makes it costly to ignore known defects.

**(5) Logged disclosure exceptions.** For any restriction applied under the disclosure ladder (L1–L3), the committee records what is withheld, why, who can access it, and a re-evaluation date or condition. These logs are public even when the substantive restricted content is not. This prevents "infohazard" from becoming an unattributable discretion.

**(6) Red-team report minimum fields.** Red-team reports are required to include a structured set of minimum fields: threat model, attack constraints, success criteria, and a public summary, even when full attack traces are controlled-access. A red-team report that omits these fields is itself a red flag under the scoring rubric.

**(7) Access-asymmetry disclosure.** Authors disclose evaluation privileges that are not available to external replicators: privileged model access, proprietary compute, non-public evaluation harnesses. This lets reviewers and downstream users weight the evidence appropriately rather than treating all claims as equally replicable.

**Goodhart's law and residual gaming risk.** We do not claim these mechanisms eliminate gaming. They are designed to beat two specific baselines. Against no checklist at all, ECAISA makes epistemic practices observable, which is a precondition for any form of accountability. Against checklist-only compliance — the regime most vulnerable to Goodhart effects — the inclusion of adversarial verification (mechanisms 1–2) and structured uncertainty registers (P7) shifts the compliance target from easily fabricated formal outputs to substantive epistemic practices that are harder to fake convincingly. An uncertainty register that lists no failure modes, for instance, is itself a red flag under P7's scoring criteria and would be flagged by a competent spot-auditor.

The residual risk is that well-resourced actors may invest in producing convincing but hollow compliance artefacts. We acknowledge this possibility and note two mitigations. First, the lottery-based independent verification (mechanism 2) means that the expected cost of fabrication scales with the probability of selection, which venues can calibrate upward for T3 claims. Second, the public correction channel (mechanism 4) creates an ongoing reputational surface: any researcher can file a public issue on a claimed artefact, creating a cost for authors whose compliance turns out to be hollow. These mechanisms do not guarantee substantive compliance, but they make non-compliance observable and costly, which is the realistic target for any governance instrument operating in a competitive research environment.

## 7.2 Adoption dynamics and the coordination problem

Voluntary adoption of ECAISA by individual laboratories runs into a structural problem. Unilateral adoption imposes costs — disclosure overhead, adversarial verification, documentation, the coordination time to run infohazard reviews — without reciprocal benefits when competitors do not adopt. Under the current deployment economics of frontier AI, where speed-to-capability confers strategic advantage in funding, hiring, and market position, a laboratory that unilaterally accepts hostile replication overhead for safety-relevant claims faces a first-mover disadvantage relative to competitors who do not. This is a classic coordination failure, and any implementation outlook that ignores it underestimates the gap between normative appeal and actual adoption.

Three realistic adoption paths reduce or bypass this unilateral disadvantage by shifting the payoffs institutionally rather than asking individual labs to accept the cost alone.

**(a) Venue-level mandates.** Venues that host safety-relevant tracks can introduce artefact-based review criteria tied to the ECAISA rubric for T2 and T3 claims: full replication packages, adversarial verification reports, and uncertainty registers become prerequisites for acceptance rather than optional supplements. This flips the payoff structure: non-compliance becomes a rejection cost rather than a competitive advantage, and the marginal cost of compliance is paid by everyone who wants the venue's imprimatur. Analogues include the CONSORT guidelines for clinical-trial reporting and the preregistration requirements at registered-report venues; neither imposes compliance as a duty, but both make it strictly worthwhile by tying it to publication access.

**(b) Funder mandates.** Funding bodies that support safety-relevant research can earmark a fraction of their safety-research budgets (we suggest in the range of 5–10%) for adversarial verification grants, awarded competitively to teams proposing credible refutation protocols for high-profile claims. This socialises the cost of independent verification across the funding pool rather than asking each lab to find resources for hostile replication of its competitors' work. Similar arrangements have made external replication programmes possible in experimental psychology and experimental economics.

**(c) Regulatory anchoring.** Regulators and standards bodies can embed ECAISA-style hooks in the procedural layer of existing regulation: explicit research-phase evidentiary requirements in EU AI Act codes of practice, ISO/IEC technical reports on AI research quality management, and procurement frameworks that make compliance legible to downstream buyers. Once present at the procurement layer, compliance becomes a competitive asset rather than a burden, and labs that adopt ECAISA unilaterally recover the cost through the procurement channel.

None of these paths eliminates first-mover disadvantage for a laboratory competing purely on capability. What they do is reduce the space in which pure capability competition remains rational for safety-relevant claims specifically. The realistic adoption trajectory is therefore neither unilateral voluntary uptake (which the coordination problem undermines) nor top-down mandate (which regulatory processes are too slow to produce in time), but a staged institutionalisation through the three paths above, beginning with venue-level artefact requirements for T2 and T3 claims and funder support for independent adversarial verification.

**Equity as infrastructure.** A complementary consideration is equity. ECAISA increases documentation and verification overhead; without institutional support, these demands risk entrenching incumbents with privileged access to compute, evaluators, and proprietary systems while excluding under-resourced researchers. Adoption should therefore be paired with concrete support: earmarked replication funding and bounties for independent reproduction and adversarial verification of T2 and T3 claims; shared, community-governed evaluation harnesses with standardised reporting; venue-provided artefact shepherding to help authors reach the P5 minimum without bespoke engineering capacity; and public-interest compute credits for reproductions and red-teaming, prioritised for external validators. These measures treat verification capacity as governance infrastructure rather than as an unfunded mandate on individual authors, and they help to break the coordination failure from the bottom up as well as the top down.

Moving forward, we outline several steps to promote adoption of these epistemic norms and to evaluate their effectiveness:

- **Pilot Projects and Case Studies:** We encourage the community to trial the full P1–P8 stack on a few high-profile alignment research projects. For example, a large team working on a deceptive alignment benchmark or advanced AI sandbox experiment could formally adopt ECAISA for the project's duration and report on outcomes. Publishing detailed meta-analyses of these trials including benefits (e.g. errors caught, improvements in collaboration) and costs (extra time, resource needs) will build an empirical case for or against the code's recommendations. Demonstrating real-world feasibility is key to broader buy-in. These pilots can also serve as exemplars or templates that others can emulate.

- **Incentives via Certification and Publication Standards:** We propose working with AI conferences, journals, and funding bodies to integrate ECAISA principles into their requirements. For instance, much like psychology journals now have “Transparency and Openness Promotion” (TOP) badges, alignment venues could establish an “Alignment Epistemic Rigor” badge or checklist. A paper that adheres to all or most of P1–P8 (preregistered, all data open, adversarial review included, etc.) might be recognized formally, giving authors incentive to do the extra work. Funding agencies could similarly prioritize grants that commit to these practices (some grant programs already ask for reproducibility plans, this would be a logical extension). Additionally, an independent body could certify labs or projects that follow the code (analogous to an ISO certification but for research process), which would signal credibility to outsiders. Over time, if such certifications or norms become prestigious, they create a race to the top on rigor, countering the current race-to-publish incentives.
- **Integration into Governance and Policy:** We will also advocate for these research-phase norms to be reflected in AI governance frameworks. Regulatory and standards bodies could incorporate ECAISA-like language into AI risk management guidelines. For example, research-phase audits or verification standards could be appended to the EU AI Act’s codes of practice, ensuring that regulators explicitly value epistemic rigor alongside technical safety. ISO/IEC could produce a technical report on “AI research quality management” referencing these principles. By getting alignment epistemics on the policy radar, we not only legitimize the effort but might eventually require through law or funding mandates that safety-critical AI research follows stricter protocols. This would amplify the impact beyond what voluntary adoption can achieve.

Ultimately, formalizing epistemic norms is itself an experiment and an iterative process. We must apply the same scrutiny to these principles that we demand for alignment claims. It is possible there are unintended consequences: for instance, could requiring preregistration and extreme proof slow down innovation or deter exploratory research? Might an overemphasis on process create bureaucratic hurdles without proportionate benefit? These are valid concerns that need empirical study.

We envision feedback loops to refine the code: monitoring the field’s productivity, creativity, and risk posture as norms change, and gathering community feedback regularly. For example, a meta-research study in a couple of years could compare alignment papers published under ECAISA guidelines vs. those that weren’t, to see if differences in replicability or error rates emerge. If we find any of the principles are too onerous or not effective, we will adjust them. In this sense, ECAISA is a starting framework grounded in timeless scientific virtues (honesty, transparency, scepticism) but also adapting to the urgent specifics of alignment (adversarial uncertainty, extreme tail risk).

We also suggest exploring additional dimensions the community raised that lie outside the scope of this paper. One is epistemic security protecting the integrity of the research process from intentional manipulation (e.g. by malicious actors spreading misinformation or by AI systems themselves influencing our beliefs). Our principles indirectly help (because openness and verification make it harder to insert false claims), but direct measures (like secure model evaluation environments or adversarial information hygiene) might be needed.

Another area is cross-disciplinary fertilization: collaborating with fields like cognitive psychology (to understand and mitigate researcher biases) or economics of science (to design incentives) could enrich the epistemic toolkit for alignment. Finally, while we have focused on research-phase norms, bridging the gap to deployment is important; how do these norms inform the transition from lab findings to real-world AI system governance? Future work could develop guidelines for how alignment research evidence should be translated into safety assurances for deployment (perhaps akin to medical translation from clinical trials to practice guidelines).

We argue that the adoption of an Epistemic Code not as a one-off fix but as the beginning of a more reflective, self-regulating phase in the evolution of AI safety science. The community's willingness to experiment on itself (i.e. to critically evaluate and improve its own epistemic practices) will be a healthy indicator that we take the notion of extreme rigor seriously.

# 8. Conclusion

AI alignment is not merely a technical puzzle; it is fundamentally an epistemic challenge: how to extend reliable knowledge and guarantees into unprecedented, high-stakes domains. The stakes, potentially the future of human civilization demand research methods that are up to this task. We have argued that the alignment community should adopt a formal Epistemic Code encompassing transparency, integrity, evidence proportionality, reproducibility, stewardship of research artifacts, organized scepticism, reflexive uncertainty accounting, and epistemic hygiene. By adhering to these principles, researchers will document everything, challenge everything, and collaborate openly. In effect, we create a culture where any claim about safety is built on a mountain of openly scrutinized evidence and where doubt is systematically welcomed to test claims.

If the optimistic scenario materialises i.e. advanced AI integrates seamlessly without inflicting harm, the quiet hero will be the discipline of our epistemic safeguards. As public-health experts know, when a pandemic never erupts, it is the unglamorous vaccination programmes and surveillance protocols that deserve the credit. In safety-critical domains, non-events are evidence that precautions worked.

The same epistemic rigour refined for AI alignment could become a template for other fields operating under radical uncertainty such as climate geo-engineering, synthetic-biology containment, even planetary-scale cybersecurity. By showing how to prosecute science when the downside risk is existential and the evidence sparse, alignment research can pioneer a broader paradigm of "high-uncertainty science."

In the final analysis, pursuing alignment is about expanding human knowledge under uncertainty but doing so responsibly. By internalising rigorous epistemic norms, the field maximises its credibility and its chances of truly solving the problem it addresses. The Alignment Epistemic Code offered here is a community-driven invitation to raise our own standards, to ensure that as we race to build safe AI, we are not cutting corners in understanding what "safe" truly means. We believe such self-imposed discipline is not a hindrance but rather the foundation for trustworthy and lasting progress toward a safer AI future.

# Acknowledgements

The author likes to thank the editor and the anonymous reviewers for their detailed, constructive, and generous feedback throughout the review process. Their comments materially improved the clarity, scope, and evidential grounding of the paper.

# Appendix A: Rubric audit codebook

This appendix specifies the coding protocol used to apply the rubric summarised in Table 4. It is intended to make the rubric reproducible, to clarify the evidentiary threshold for each score, and to specify conservative decision rules for ambiguous cases.

## A.1 Unit of analysis and allowed evidence sources

Unit of analysis. One "paper" consists of the main manuscript together with any officially linked appendices and supplementary materials, and any author-linked artefacts explicitly referenced in the paper at the time of coding.

Allowed evidence sources. Coders consider only: (i) the PDF text (including appendices), (ii) links embedded in the paper (URLs, DOIs), and (iii) content reachable via those links without guessing additional resources. Unpublished materials are not considered, practices are not inferred from author reputation, and informal claims ("available on request") are not treated as evidence unless an access pathway is concretely specified.

Conservative principle. If evidence for a criterion is ambiguous, missing, or not verifiable from the allowed sources, the lower score is assigned and the ambiguity is recorded in the justification field.

## A.2 Scoring scale and general decision rules

Each principle P1–P8 is coded on an ordinal scale: 0 (absent) = no substantive evidence of the practice; 1 (partial) = some evidence or partial implementation, but falling short of the operational threshold for "meets"; 2 (meets) = satisfies the operational threshold defined in A.3 below. Tie-break rule: when deciding between 1 and 2, prefer 1 unless the "meets" criterion is explicitly evidenced; when deciding between 0 and 1, prefer 0 unless substantive evidence of the practice is present beyond generic prose.

## A.3 Per-principle criteria

**P1 — Transparency and disclosure.** Score 0: no code, data, or evaluation details beyond prose; no stable links; no artefact access plan. Score 1: some artefacts provided or methods described plausibly supporting reimplementation, but disclosure is incomplete or links are unstable. Score 2: sufficient artefacts released to enable meaningful scrutiny, or a concrete disclosure protocol applied under the ladder (Table 5): (i) explicit list of withheld items, (ii) justification, (iii) mitigations and access pathway, and (iv) a re-evaluation condition or date.

**P2a — Selective-reporting prevention.** Score 0: selective reporting evident; negative results and nulls absent; no limitations discussion. Score 1: some limitations noted but no demonstrated commitment to comprehensive reporting. Score 2: explicit anti-selection practices including (i) failures and nulls reported alongside successes, (ii) sufficient detail for others to challenge results, and (iii) pre-specified primary outcomes or a clear post-hoc rationale for reported metrics.

**P2b — Conflict-of-interest and access disclosure.** Score 0: no funding or COI disclosure; no acknowledgement of access asymmetries. Score 1: basic funding or COI disclosed but privileged access affecting replicability not noted. Score 2: full disclosure: (i) all funding and relevant conflicts named, and (ii) any access asymmetries an independent replicator would not share are explicitly stated.

**P2c — Provenance and error-correction integrity.** Score 0: no mechanism for post-publication correction; no versioning or lifecycle logging. Score 1: some artefact versioning exists but no explicit correction channel or errata policy. Score 2: operational correction infrastructure: (i) public issue tracker or errata page with stated update policy; and (ii) for T3 claims, tamper-evident provenance for core lifecycle events where feasible.

**P3 — Pre-specification and quantified evaluation.** Score 0: largely qualitative claims without clear metrics or baselines. Score 1: quantitative evaluation exists but under-specified (unclear baselines, limited robustness checks). Score 2: measurable success criteria with (i) clear metrics and baselines aligned to claims, (ii) sufficient experimental detail, and (iii) at least one robustness-oriented element proportionate to the claim.

**P4 — Independent and adversarial verification.** Score 0: no replication or independent verification. Score 1: strong internal verification but no independent replication or adversarial audit documented. Score 2: central claim subjected to an independent check (one of): independent replication by a separate team; structured third-party red-teaming with reportable traces; or a release package designed for adversarial verification plus explicit invitation and support for external replication, with evidence of uptake or a clearly defined protocol.

**P5 — Artefact stewardship.** Score 0: no artefacts or only ephemeral links. Score 1: artefacts exist but lack stewardship features (no version tags, no stable archival link or DOI, no environment spec). Score 2: artefacts stewarded as citable, durable research objects: (i) versioned release, (ii) stable identifier, (iii) environment specification, (iv) licence and provenance notes, and (v) maintenance or ownership statement.

**P6 — Red-teaming and adversarial evaluation.** Score 0: no explicit adversarial or misuse-oriented testing. Score 1: some adversarial testing but without a clear threat model, systematic method, or reproducible traces. Score 2: structured adversarial evaluation: (i) threat model and attack surface; (ii) systematic test design and constraints; (iii) clear reporting of successes and failures; and (iv) reproducible artefacts or documented exceptions.

**P7 — Uncertainty accounting and calibration.** Score 0: no meaningful uncertainty accounting beyond generic limitations. Score 1: uncertainties discussed in a non-structured way but lacks an explicit register or decision-relevant calibration. Score 2: explicit uncertainty artefacts including at least (i) a structured uncertainty register (assumptions, known unknowns, failure modes), (ii) a residual-risk statement and plausible falsifiers, and (iii) calibration or sensitivity analyses where applicable.

**P8 — Institutional positioning and decision relevance.** Score 0: no discussion of how claims should be used or misused; no COI disclosure; no governance context. Score 1: basic disclosure and some normative positioning but no operational guidance for how the result should inform decisions. Score 2: claims explicitly situated in an institutional decision context: (i) clear claim boundaries ("what this does and does not imply"), (ii) explicit COI, funding, and relevant access-asymmetry disclosure, and (iii) concrete decision-facing guidance.

### A.4 Inter-rater reliability

For the pilot application of this codebook, a subset of papers is double-coded by an independent coder after a brief calibration exercise on a paper outside the audit. We report percent agreement across coded cells and weighted Cohen's κ with quadratic weights for the ordinal 0/1/2 scale (Cohen 1960, 1968). Disagreements are adjudicated using the decision rules above, with the default rule "prefer the lower score unless the criterion is explicitly evidenced." Reliability indicators from a pilot overlap should be interpreted as evidence of scorability, not as a mature psychometric validation of the instrument; we recommend expanding double-coding in any larger subsequent study.

### A.5 Coding workflow and justification requirements

Workflow: (1) read the abstract, introduction, and contributions to identify what the paper asserts; (2) identify author-linked artefacts and record accessibility and coding date; (3) code P1–P8 using the criteria above, recording a short justification per code; (4) record any ambiguous case and the reason for conservative scoring. The recommended annotation sheet fields are: paper ID, venue and year, claim type, scores P1–P8, justifications per principle, coding date, and artefact-access notes.

## Appendix B: Coverage-matrix justification

This appendix substantiates every filled cell in the coverage matrix (Figure 1) by stating the specific obligation each principle imposes that closes the relevant dimension's failure mode. The appendix is intended to convert the matrix from an asserted mapping into an auditable claim: for each (principle, dimension) pair where a coverage mark appears, Table 10 names the concrete epistemic failure being addressed and the obligation through which the principle addresses it. Cells not listed are intentionally empty in the matrix.

**Table 10. Coverage-matrix justification. Each row lists one filled cell in Table 3 and names the failure mode it addresses and the obligation imposed.**

| Principle | Dimension | Failure mode addressed | Obligation imposed by the principle |
|---|---|---|---|
| P1 | E1 Optimisation target | Worst-case evaluations hidden behind average-case reporting. | Preregister hypotheses and release evaluation pipelines so worst-case tests cannot be silently dropped. |
| P1 | E2 Transparency | Delayed or partial disclosure allows hidden failure modes to persist. | Release code, data, and checkpoints at study outset in FAIR repositories (subject to the disclosure ladder). |
| P1 | E5 Enforcement culture | Delayed disclosure weakens external review. | Public preregistration creates an auditable commitment that reviewers can check against. |
| P2 | E2 Transparency | Selective reporting and undeclared interests distort the risk record. | Require reporting of nulls, COI disclosure, access-asymmetry disclosure, and tamper-evident provenance. |
| P2 | E3 Verification | Missing provenance and undisclosed access make independent verification impossible. | Sub-obligation P2c: public correction channel with stated errata policy; P2b access-asymmetry disclosure. |
| P2 | E5 Enforcement culture | Selective reporting and hidden COIs normalise low-integrity practice. | Publicly log reporting commitments and COI; sanction departures via the correction channel. |
| P3 | E1 Optimisation target | Claim strength untethered from the stringency of tail-risk evidence. | Bind claim strength to quantified support; preregister analysis plans; require robustness checks proportionate to stakes. |
| P3 | E2 Transparency | Overclaiming obscures the real evidential basis. | Transparent metrics and baselines; explicit absence-of-evidence caveats. |
| P3 | E3 Verification | Weak effect sizes inflated into strong claims without robustness. | At least one robustness-oriented element proportionate to claim strength. |
| P3 | E4 Uncertainty | Overclaiming rooted in under-quantified uncertainty. | Require explicit uncertainty range alongside any headline effect size. |
| P4 | E1 Optimisation target | Worst-case claims unchallenged by adversaries. | Mandate at least one independent adversarial verification for major safety claims. |
| P4 | E3 Verification | Single-team results ossify without independent replication. | Full replication package; structured invitation for external replicators. |
| P4 | E5 Enforcement culture | Lack of replication incentives reinforces single-team epistemics. | Venues and funders are asked to reward adversarial verification as first-class work (see Implementation Outlook). |
| P5 | E2 Transparency | Ephemeral artefacts cannot be audited after publication. | DOIs, versioned releases, rich metadata, and long-term maintenance statements. |
| P5 | E3 Verification | Non-reproducible environments block replication attempts. | Environment spec (lockfile or container), hash manifests for critical files. |
| P5 | E5 Enforcement culture | Lack of shared artefact standards makes external scrutiny costly. | Community-governed stewardship with reproducible build pipelines for T3 claims. |
| P6 | E3 Verification | Adversarial review is informal | Make red-team review the default for |

| | | and unincentivised. | safety claims; publish critique reports as first-class contributions. |
|---|---|---|---|
| P6 | E5 Enforcement culture | Weak institutional scepticism allows normalisation of deviance. | Separation-of-concerns in internal safety teams; career credit and bounties for flaw-finding. |
| P7 | E4 Uncertainty | Shallow uncertainty reporting; epistemic unknowns go unrecorded. | Maintain a living uncertainty register with assumptions, failure modes, falsifiers, and residual-risk statements. |
| P7 | E1 Optimisation target | Tail-risk claims made without quantifying what is unknown. | Require explicit residual-risk quantification aligned to the tail-risk target. |
| P8 | E5 Enforcement culture | Research used as compliance proxy outside its intended scope. | State claim boundaries; give decision-facing guidance with reliance conditions and update triggers. |